\documentclass[11pt]{article}
\usepackage[margin=1in]{geometry}
\usepackage[T1]{fontenc}
\usepackage[utf8]{inputenc}
\usepackage{lmodern}
\usepackage{microtype}
\usepackage{amsmath,amssymb,amsthm}
\usepackage{mathtools}
\usepackage{siunitx}
\usepackage{graphicx}
\usepackage{booktabs}
\usepackage{natbib}
\usepackage{enumitem}
\usepackage{hyperref}
\usepackage[capitalize,nameinlink]{cleveref}
\usepackage{algorithm}
\usepackage{algpseudocode}
\usepackage{xcolor}
\usepackage{tikz}
\usepackage{placeins}

\hypersetup{colorlinks=true,linkcolor=blue,citecolor=teal,urlcolor=magenta}

\crefname{algorithm}{Alg.}{Algs.}
\Crefname{algorithm}{Algorithm}{Algorithms}

\newtheorem{definition}{Definition}
\newtheorem{proposition}{Proposition}

\newcommand{\ERT}{\operatorname{ERT}}

\title{Time-Fair Benchmarking for Metaheuristics: A Restart-Fair Protocol for Fixed-Time Comparisons}
\author{Junbo Jacob Lian\\\small Northwestern University \\\small \texttt{JacobLian@u.northwestern.edu}}
\date{September 9, 2025}

\begin{document}
\maketitle

\begin{abstract}
Numerous purportedly ``improved'' metaheuristics claim superior performance based on equivalent function evaluations (FEs), yet often conceal additional computational burdens in more intensive iterations, preprocessing stages, or hyperparameter tuning. This paper posits that \textbf{wall-clock time}, rather than solely FEs, should serve as the principal budgetary constraint for equitable comparisons. We formalize a fixed-time, restart-fair benchmarking protocol wherein each algorithm is allotted an identical wall-clock time budget per problem instance, permitting unrestricted utilization of restarts, early termination criteria, and internal adaptive mechanisms. We advocate for the adoption of anytime performance curves, expected running time (ERT) metrics to predefined targets, and performance profiles employing time as the cost measure. Furthermore, we introduce a concise, reproducible checklist to standardize reporting practices and mitigate undisclosed computational overheads. This approach fosters more credible and practically relevant evaluations of metaheuristic algorithms.
\end{abstract}


\section{Motivation}

In the domain of metaheuristic optimization, comparative evaluations frequently rely on a fixed budget of function evaluations (FEs) to assess algorithmic variants. However, this metric can be misleading due to substantial disparities in per-iteration computational costs arising from elements such as surrogate modeling, extensive neighborhood explorations, intricate population dynamics, or parallel processing implementations. Community surveys and best-practice guidelines warn that FE-only reporting may bias conclusions and harm reproducibility \citep{BartzBeielstein2020,Sala2020CEC}. For instance, a sophisticated variant incorporating machine learning-based surrogates may incur significantly higher overhead per iteration compared to a simpler baseline, rendering FE-equivalent comparisons inherently biased.

Consider a hypothetical scenario where a proposed advanced particle swarm optimization (APSO) requires \SI{50}{s} for a single execution, while a standard PSO baseline completes in \SI{10}{s}. Under traditional FE-based benchmarking, both might be allocated the same number of evaluations, potentially favoring the APSO if its enhancements yield better solutions within that FE limit. Yet, in practical applications, users are constrained by wall-clock time rather than abstract evaluation counts. A time-fair comparison would permit the baseline PSO to execute multiple independent restarts within the \SI{50}{s} budget, selecting the optimal solution among them. This perspective aligns benchmarking with real-world decision-making: what is the best attainable solution quality within a specified temporal allocation?

Moreover, undisclosed costs in preprocessing, such as initial population generation or feature selection, and hyperparameter tuning can skew results. Reviews highlight that many ``novel'' metaheuristics overlook these aspects, leading to inflated performance claims \citep{BartzBeielstein2020,Schott2021}. Recent optimizers have begun incorporating early stopping and reinitialization strategies to enhance efficiency and escape local optima within fixed budgets \citep{Tang2025}. By emphasizing wall-clock time, our protocol encourages transparency and incentivizes efficient algorithmic designs that balance exploration, exploitation, and computational efficiency.

\section{Protocol}
We delineate a streamlined, algorithm-agnostic protocol applicable to both continuous and discrete black-box optimization paradigms. This framework prioritizes wall-clock time as the unifying budget, accommodating diverse internal strategies while ensuring comparability.
\begin{definition}[Fixed-Time Budget]
Let $\mathcal{I}$ denote a collection of problem instances, and let $T > 0$ represent the wall-clock time budget allocated per instance. An algorithm $\mathcal{A}$ is permitted to employ any amalgamation of restarts, internal parallelism, adaptive mechanisms, and early stopping heuristics, provided the aggregate execution time remains within $T$.
\end{definition}
\begin{definition}[Restart-Fairness]
For a stipulated $T$, an algorithm exhibiting an average single-run duration $\tau$ may execute $k = \lfloor T / \tau \rfloor$ independent runs, reporting the optimal objective value across these executions. Comparative baselines are afforded equivalent latitude in restart utilization.
\end{definition}
\begin{definition}[Time-to-Target and ERT]
Given a quality threshold $q$, denote by $t_i$ the time elapsed to attain $q$ in the $i$-th run (truncated at $T$ if unsuccessful). With $s$ successful runs out of $R$ total attempts, the expected running time is
\begin{equation}
\label{eq:ert}
\ERT(q) = \frac{\sum_{i=1}^{R} \min\{t_i, T\}}{s} , \quad s \ge 1.
\end{equation}
In cases where $s = 0$, declare $\ERT(q) = \infty$ and furnish the empirical success rate $s/R$.
\end{definition}
\paragraph{Anytime Performance Curves.}
Under a fixed budget $T$, chronicle the evolution of the best-so-far objective value as a function of elapsed time. For aggregation across instances, employ empirical cumulative distribution functions (ECDFs) or median trajectories augmented with confidence intervals, facilitating visual assessment of convergence dynamics \citep{LopezIbanez2010EAF}.
\paragraph{Performance Profiles with Time as Cost.}
Adapting the methodology of \citet{Dolan2002}, define the cost metric for each solver-instance pair as, inter alia, the ERT to a predetermined tolerance or the time to achieve a fixed quality level. Subsequently, plot the proportion of instances for which a solver's cost lies within a factor $\tau$ of the minimal cost observed across all solvers.
\paragraph{Accounting for Tuning and Ancillary Costs.}
Preprocessing and hyperparameter optimization expenditures must be explicitly reported. Where practicable, amortize these costs over the instance set \citep{Hutter2019AlgConf}. Delineate hardware specifications (e.g., CPU/GPU models, core counts), threading paradigms, linear algebra libraries (e.g., BLAS), compiler directives, and software versions to ensure reproducibility.
\begin{proposition}
Under the fixed-time restart-fair protocol, algorithms with lower per-run overhead are incentivized to leverage multiple restarts, potentially mitigating local optima entrapment and enhancing overall solution quality, as supported by restart strategy analyses \citep{Kadioglu2017,Friedrich2017BetAndRun,Weise2018,Tang2025}.
\end{proposition}

\section{Illustrative PSO Example}

To elucidate the protocol, consider a hypothetical comparison between a proposed adaptive PSO (APSO) and a standard PSO on a benchmark function, such as the Rastrigin function in 10 dimensions. Assume APSO incorporates surrogate-assisted updates, resulting in a single-run duration of \SI{50}{s}, whereas standard PSO completes in \SI{10}{s}. Note that this APSO is a fictional construct for illustrative purposes, and no specific literature describes such an algorithm exactly as presented here.

In a traditional FE-based evaluation with a budget of 10{,}000 evaluations, APSO might achieve a mean best fitness of 5.2, outperforming PSO's 8.7 due to its advanced mechanisms. However, under our time-fair protocol with $T = \SI{50}{s}$, PSO can perform up to five independent restarts. Aggregating the best solution across restarts, PSO attains a mean fitness of 4.1, surpassing APSO's single-run performance.

Further, employing time-to-target metrics with a target fitness of 5.0, PSO exhibits an average time-to-target of \SI{18}{s} per successful run, yielding $\ERT = \SI{20.5}{s}$ across 20 repetitions (success rate 95\%). APSO, with \SI{22}{s} average time-to-target and 85\% success, results in $\ERT = \SI{25.9}{s}$. This inversion underscores how time-fairness reveals practical efficacies obscured by FE equivalence.

In practice, restart strategies can be adaptive; for instance, monitoring stagnation via fitness plateaus to trigger restarts \citep{Kadioglu2017}. Such mechanisms are seamlessly integrated within our protocol, as exemplified in recent optimizers that employ early stopping and reinitialization to manage computational budgets effectively \citep{Tang2025}.

\section{Reporting Checklist}
To promote standardization and transparency, we advocate inclusion of the following elements in primary manuscripts or supplementary materials:

\begin{enumerate}[leftmargin=*]
\item \textbf{Budget Specification:} Declare the fixed wall-clock time $T$ per instance; supplement with internal FE caps if applicable.
\item \textbf{Restart Policies:} Affirm allowance of restarts for all algorithms; report average number of completed runs within $T$.
\item \textbf{Target Definitions:} Specify absolute or relative quality thresholds; articulate success criteria.
\item \textbf{Performance Metrics:} Present anytime curves, ERT to targets, and time-based performance profiles.
\item \textbf{Statistical Rigor:} Detail random seeds, repetitions per instance; incorporate confidence intervals or nonparametric statistical tests.
\item \textbf{Computational Environment:} Enumerate hardware (CPU/GPU models, cores/threads), software libraries, versions, and optimization flags.
\item \textbf{Tuning Overhead:} Quantify time and methodology for hyperparameter optimization; indicate amortization strategies.
\item \textbf{Reproducibility Artifacts:} Provide source code, seeds, and scripts for figure generation and result replication.
\end{enumerate}

Adherence to this checklist mitigates common pitfalls in metaheuristic benchmarking, as identified in recent surveys \citep{BartzBeielstein2020,Schott2021}.

\section{Figures (placeholders)}
The ensuing placeholders employ TikZ for compilation sans external data; in actual submissions, supplant with empirical plots derived from benchmark suites like BBOB \citep{Hansen2010BBOBsetup} or IOHprofiler \citep{Doerr2018IOHprofiler}.

\begin{figure}[h]
\centering
\begin{tikzpicture}[x=1cm,y=1cm]
\draw[thick] (0,0) rectangle (12,5);
\node at (6,4.5) {\small \textbf{Anytime curve placeholder}};
\draw (1,1) -- (5,3.2) -- (9,4.2) -- (11,4.5);
\draw[dashed] (1,0.8) -- (5,2.0) -- (9,3.0) -- (11,3.6);
\node at (7.8,1.0) {time};
\node[rotate=90] at (-0.6,2.5) {best-so-far $f$};
\end{tikzpicture}
\caption{Anytime performance trajectories under fixed time $T$. Solid line: baseline algorithm with multiple restarts. Dashed line: proposed variant with higher per-iteration cost.}
\label{fig:anytime}
\end{figure}
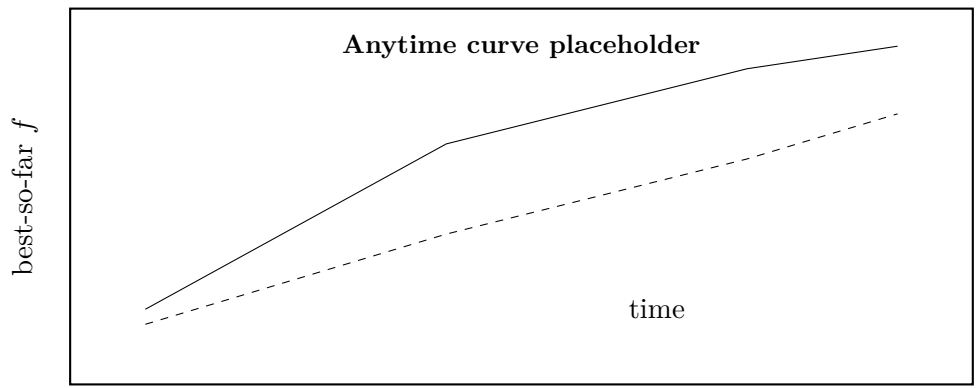

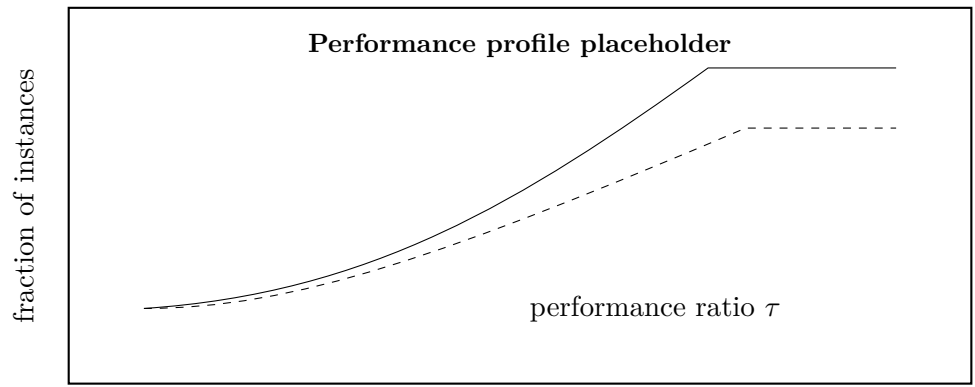
\begin{figure}[h]
\centering
\begin{tikzpicture}[x=1cm,y=1cm]
\draw[thick] (0,0) rectangle (12,5);
\node at (6,4.5) {\small \textbf{Performance profile placeholder}};
\draw (1,1) .. controls (4,1.2) and (6,2.4) .. (8.5,4.2) -- (11,4.2);
\draw[dashed] (1,1) .. controls (3,1.0) and (5,1.7) .. (9,3.4) -- (11,3.4);
\node at (7.8,1.0) {performance ratio $\tau$};
\node[rotate=90] at (-0.6,2.5) {fraction of instances};
\end{tikzpicture}
\caption{Performance profiles utilizing \emph{time} (e.g., ERT or time-to-target) as the underlying cost metric.}
\label{fig:pp}
\end{figure}

\section{Related Efforts}
Performance profiles \citep{Dolan2002} offer a robust framework for synthesizing comparative solver performances across diverse instance sets. The COCO/BBOB initiative \citep{Hansen2010BBOBsetup} champions time/FE-to-target evaluations, anytime ECDF aggregations, and rigorous statistical methodologies. Complementary tools like IOHprofiler \citep{Doerr2018IOHprofiler} automate benchmarking pipelines and visualizations, while exploratory analyses \citep{LopezIbanez2010EAF} establish best practices for ECDF and empirical attainment function (EAF) curves in stochastic settings.

Restart strategies have been extensively studied, with adaptive and learning-based approaches demonstrating efficacy in escaping local optima \citep{Kadioglu2017,Weise2018,Friedrich2017BetAndRun}. Algorithm configuration techniques \citep{Hutter2019AlgConf} address tuning costs, aligning with our emphasis on amortization. Benchmarking tools tailored for metaheuristics \citep{Schott2021} introduce composite efficiency scores incorporating convergence speed and solution quality, resonating with our time-centric metrics. Furthermore, novel optimizers like IECO incorporate early stopping and reinitialization within fixed evaluation budgets, providing practical inspiration for restart-fair protocols \citep{Tang2025}.

Our protocol synthesizes these elements into a cohesive, restart-fair framework that privileges wall-clock time, explicitly mitigates hidden costs, and promotes reproducible practices, addressing open challenges in metaheuristic benchmarking \citep{BartzBeielstein2020,Sala2020CEC}.

\section{Limitations and Scope}
This protocol is primarily tailored for single-objective black-box optimization scenarios where algorithmic overhead is non-negligible relative to evaluation times. In contexts where function evaluations dominate computational costs (e.g., expensive simulations), FE-based budgets may remain appropriate; nonetheless, reporting wall-clock times provides supplementary insights, particularly under heterogeneous hardware or parallelization schemes. Extensions to multi-objective or dynamic optimization warrant future investigation.

\section{Conclusion}
By anchoring comparisons in wall-clock time and endorsing restart-fair practices, our protocol bridges the gap between academic evaluations and practical deployments, where temporal constraints predominate. Widespread adoption should curtail spurious superiority claims predicated on FE parity alone, fostering verifiable advancements in metaheuristic design.

\paragraph{Reproducibility Statement.} Accompanying scripts implementing \Cref{alg:protocol,alg:restart}, along with utilities for generating anytime curves, ERT computations, and performance profiles from execution logs, will be publicly disseminated.

\begin{algorithm}[t]
\caption{Time-Fair Evaluation Protocol (per instance)}\label{alg:protocol}
\begin{algorithmic}[1]
\Require time budget $T$, target set $\mathcal{Q}$ (optional), repetitions $R$ per algorithm
\For{each algorithm $\mathcal{A}$}
\State $\text{time\_used} \gets 0$, $\mathcal{S} \gets \emptyset$ \Comment{store independent run summaries}
\While{$\text{time\_used} < T$}
\State commence timer; execute $\mathcal{A}$ once with independent seed until termination or early success
\State log best-so-far trajectory and time-to-targets for $\mathcal{Q}$
\State halt timer; append run summary to $\mathcal{S}$; increment $\text{time\_used}$
\EndWhile
\State Aggregate: best-of-restarts at $T$; compute ERT for each target in $\mathcal{Q}$
\EndFor
\State Across instances: construct anytime ECDFs and time-based performance profiles.
\end{algorithmic}
\end{algorithm}

\begin{algorithm}[t]
\caption{Baseline with Independent Restarts (e.g., PSO)}\label{alg:restart}
\begin{algorithmic}[1]
\Require budget $T$, single-run max time $\tau$, seeds $\{\sigma_j\}$
\State $k \gets \lfloor T/\tau \rfloor$, $f^* \gets +\infty$
\For{$j=1$ to $k$}
\State Execute PSO$(\sigma_j)$ for at most $\tau$ seconds; yield best value $f_j$
\State $f^* \gets \min(f^*, f_j)$
\EndFor
\State \Return $f^*$
\end{algorithmic}
\end{algorithm}

\FloatBarrier

\bibliographystyle{plainnat}
\bibliography{refs}

\end{document}